\documentclass[11pt]{article}

\usepackage[margin=1in]{geometry}
\usepackage{times}
\usepackage{cite}
\usepackage{amsmath,amssymb,amsfonts}
\usepackage{algorithm}
\usepackage{algorithmic}
\usepackage{graphicx}
\usepackage{textcomp}
\usepackage{xcolor}
\usepackage{booktabs}
\usepackage{subcaption}
\usepackage{hyperref}
\usepackage{multirow}

\title{The Illusion of Equivalence: Systematic FP16 Divergence in KV-Cached Autoregressive Inference}

\author{
Ranjith Chodavarapu\\
Department of Computer Science\\
Kent State University, USA
\and
Lei Xu \\
Department of Computer Science\\
Kent State University, USA
}

\date{}

\begin{document}

\maketitle

\begin{abstract}

KV caching is a ubiquitous optimization in autoregressive transformer inference, long presumed to be numerically equivalent to cache-free computation. This assumption fails under standard FP16 precision: cache-ON and cache-OFF execution paths employ different floating-point accumulation orderings which, due to FP16 non-associativity, produce a deterministic divergence in decoded token sequences. Across three open-weight models (LLaMA-2-7B, Mistral-7B-v0.3, Gemma-2-2B) evaluated on GSM8K, we observe a 100\% token divergence rate across all sampling strategies,  including greedy decoding, which rules out sampling randomness as a cause, and also with cache-ON yielding higher accuracy in 8 of 9 conditions, where the accuracy difference serves as an indicator that the divergence direction is systematic rather than random. Controlled FP32 falsification reduces divergence by eight orders of magnitude, eliminates token flips, and drops the flip rate to exactly 0.0\%, confirming FP16 non-associativity as the sole causal driver. Layer-wise drift profiling reveals architecturally predictable propagation patterns: models using Grouped-Query Attention exhibit sharp divergence at the first layer, while Gemma's larger head dimension and sliding window attention produce uniform accumulation across all layers. Finally, activation patching of the entire residual stream fails to recover the cache-free trajectory, localizing the causal variable to the stateful KV cache. These findings establish that FP16 KV cache inference is fundamentally non-equivalent to recomputation and provide a mechanistic framework for understanding numerical instability in modern LLM inference systems.

\end{abstract}

\textbf{Keywords} - KV cache, FP16 inference, numerical divergence, activation patching, inference reliability

\section{Introduction}
Autoregressive transformer inference is almost always based on the Key-Value (KV) cache: instead of recomputing attention over the entire prefix at every decoding step, previously computed keys and values are stored and retrieved as needed. This technique, found in practically every production serving system~\cite{kwon2023efficient}, efficient attention kernels~\cite{dao2022flashattention, dao2023flashattention}, and a prompt-caching API, is typically assumed to be numerically equivalent to cache-free inference. We show this to be untrue in the simplest case of a standard FP16 inference.

The cache-ON and cache-OFF execution paths have different layouts and kernel structures, and slightly different floating-point accumulation order (since FP16 arithmetic is not associative, $(a + b) + c \neq a + (b + c)$). These differences lead to a systematic numerical divergence in the KV tensors written during decoding.  As opposed to the stochastic variability demonstrated in previous work~\cite{zheng2023judging, Chen2023HowIC}, this divergence is deterministic and reproducible:  for the same input, same model, and same hardware, cache-ON and cache-OFF inference will always compute different output tokens.  This divergence should not be taken to be noise;  it is a characteristic of FP16 cached inference.

Although KV caching is used ubiquitously in all performance-optimized deployed transformer systems, no work has ever characterized this divergence in a systematic way, identified its cause, or found where it occurs in the model. While prior numerical methods~\cite{micikevicius2017mixed} look at training stability or model compression~\cite{frantar2022gptq, dettmers2022gpt3},  prior work on optimizing the KV cache~\cite{kwon2023efficient, shazeer2019fast, ainslie2023gqa} assumes exact equivalence without verification. Mechanistic Interpretability methods~\cite{meng2022locating, wang2022interpretability} look at the residual stream and have not studied the stateful cache as a source of divergence, which this paper now does.

We provide a systematic empirical analysis of KV cache divergence across three open-weight language models like LLaMA-2-7B, Mistral-7B-v0.3, and Gemma-2-2B, evaluated on GSM8K~\cite{cobbe2021training} under greedy, top-k, and top-p sampling. Our analysis proceeds through five experiments that follow a causal chain. We first establish that cache-ON and cache-OFF consistently produce different outputs across all conditions, then characterize how the divergence propagates layer by layer through the network. We identify FP16 accumulation order as the root cause through a controlled FP32 falsification, resolve an apparent contradiction between flip index and KL divergence through decision boundary analysis, and localize the divergence to the KV cache state rather than the residual stream via activation patching.


Our results are consistent across all three models and all three sampling strategies, and point to a single mechanistic interpretation: FP16 rounding errors are written into KV tensors at each decode step, accumulate across layers and sequence positions, and cannot be corrected by downstream residual stream interventions. Architectural choices amplify this effect: grouped-query attention (GQA) broadcasts a single FP16 rounding error to all attending query heads simultaneously, producing correlated rather than independent errors across heads and amplifying divergence relative to standard multi-head attention (MHA).

We make the following contributions:

\begin{itemize}

    \item \textbf{Behavioral characterization.}  We show that KV cache divergence produces different token sequences in 100\% of cases across all model/strategy conditions. Under our evaluation protocol, cache-ON produces higher accuracy in 8 of 9 conditions (McNemar test, Bonferroni-corrected $p < 0.05$), an indicator of systematic execution-path bias rather than a measure of absolute reasoning ability, consistent with a systematic rather than random divergence direction. 
    Effects are consistent across random seeds and correlated with decision boundary proximity.
    
    \item \textbf{Mechanistic root cause.} For the FP16 accumulation order, we use a controlled falsification to identify the causal mechanism: by switching to FP32, the KV divergence drops to a noise floor of $2.5 \times 10^{-8}$, a reduction of more than eight orders of magnitude relative to FP16 full-model accumulated KL divergence (per-step accumulation ratio is $150$--$270\times$ across models). The mean relative accumulation error is flat regardless of sequence length ($\approx 0.036\%$), confirming error magnitude is architecture-determined and input-independent.  

    \item \textbf{Decision boundary resolution.} An apparent contradiction arises between step-count and KL divergence: early token flips are associated with higher KL, seemingly contradicting the accumulation hypothesis. Both early flips and higher KL are jointly caused by model uncertainty rather than divergence magnitude, confirmed in 8 of 9 conditions ($t$-test, $p < 0.05$). This explains behaviorally when divergence is relevant and is consistent with the timing of flip predictions.
            
    \item \textbf{Causal localization.} Activation patching of residual stream hidden states results in a median recovery of less than 1\% of the KV cache divergence across all layers, with the total patching of all layers simultaneously resulting in a negative mean recovery. Patching the residual stream increases, rather than decreases, the divergence on average. This null result localizes the divergence to the KV cache state, as opposed to attention-head or MLP-layer phenomena, and establishes that correct intervention requires direct KV cache patching rather than residual stream manipulation.
    
\end{itemize}

Together, these findings show that FP16 KV cache inference is not deterministically equivalent to no-cache inference,  specify when the lack of such an equivalence is relevant in behavior, and propose a mechanism that informs robust LLM inference system design.

\section{Background}
\subsection{KV Cache and Execution Path Differences}
Autoregressive transformers employ a Key–Value (KV) cache to speed up decoding: previously computed attention keys and values for each token in the input prefix are stored and reused in later steps. cache-OFF computes the attention over the entire prefix for every token, while cache-ON attends only to cached representations of the prefix. The cache-ON/cache-OFF behaviors are mathematically equivalent,  but are implemented in a different order,  different kernel layout, and different memory layout,  and hence exhibit different floating point accumulation. This equivalence is typically just assumed but not systematically tested across model architectures and precision formats.

Additional distinctions are introduced by attention variants, like multi-query attention (MQA) ~\cite{shazeer2019fast} uses a single KV head shared across all query heads, while grouped-query attention (GQA)~\cite{ainslie2023gqa} generalises this to a user-defined sharing ratio $R$ whereby we group query heads into sets that share one KV head. They both alter the reduction structure in ways that change the propagation of FP16 rounding errors. Thus, KV caching establishes a stateful pathway through computation: small, numeric discrepancies within cached tensors are preserved, impacting every step of subsequent decoding.

\subsection{Floating-Point Precision and Accumulation Order}
As a special type of computer program, transformer inference also relies on floating point numbers, and the calculations are not associative~\cite{goldberg1991every}:
\begin{equation}
    \begin{cases}
        (a + b) + c           \neq a + (b + c)\\
        (a \times b) \times c \neq a \times (b \times c)
    \end{cases}
\end{equation}
\footnote{Addition non-associativity is the primary driver of the divergence reported here, as attention computation involves large dot-product accumulations. Multiplication non-associativity in floating-point arithmetic is a more general property~\cite{goldberg1991every} but plays a secondary role in this context.}
Compared with a high precision format like FP32,  low precision floating point numbers like FP16 are more sensitive to the computation order.
Hence, the variations in execution order – such as cache-ON versus cache-OFF, can have systematic effects. Previous research has examined reduced precision and quantization (\cite{micikevicius2017mixed},~\cite{xiao2023efficient}), but mostly for efficiency over inference correctness. In deep transformers, small initial deviations accumulate layer-by-layer, and with KV caching, persist across decode steps, producing structured divergence rather than random noise.

\subsection{From Numerical Drift to Behavioral Changes}

Differences at the hidden state level may not lead to differences at the output level.  Decisions are made on the basis of differences in log probabilities,  where a tiny change will influence the argmax only when the competing output tokens are in close competition:  when the model is on the decision boundary.

This is in effect why behavioral divergence depends not just on the size of the numerical drift but also on the logit margin structure.  It turns out an early perturbation may switch a downstream decision because the model was uncertain,  rather than because the perturbation was large.

\subsection{Grouped-Query Attention and Amplified Accumulation} 
GQA and MQA share the same set of key and value heads between multiple query heads, reducing memory bandwidth requirements. MQA is the special case where one KV head is shared across all query heads ($R = H$);  GQA is a generalisation of MQA where we can choose a sharing ratio $1 \leq R < H$.  MHA is the case when $R = 1$, in which case each query head attends to its own separate KV head.

Under cache-ON,  the FP16 rounding in each shared KV head is broadcast to all $R$ query heads that are present. This implies that one rounding error leads to $R$ correlated errors instead of $R$ independent errors, amplifying divergence relative to MHA. The larger $R$ is, the stronger this amplification. Hence, the fact that   Mistral (GQA ratio 4:1, $R=4$) exhibits systematically higher per-layer drift than LLaMA-2 (MHA, $R=1$), and that MQA models would be expected to experience the strongest impact of all.


\subsection{Mechanistic Interpretability and Intervention Limits}
Mechanistic Interpretability approaches examine the causal architecture of transformers by intervening on internal activations.  Methods like causal tracing and ROME ~\cite{meng2022locating} and path patching~\cite{wang2022interpretability} generally work within the residual stream, which accumulates computation over the layers.

If patching residual states fails to recover model behavior,  then the causal variable is not in the residual stream; it could, at most, be in the KV cache, which contains information that is not accessible by residual stream interventions. This offers a theoretically justified way of localizing divergence to the KV cache state, as opposed to one or more specific layers or attention heads.

\subsection{Determinism and Reliability in LLM Inference}
Even with deterministic decoding methods, such as greedy sampling, transformer outputs are known to be sensitive to floating-point nondeterminism, parallel reduction order, hardware-specific kernels, and more. Examples of previous work showing such output instability and reproducibility issues with large LLMs include: surveys of hallucination and inconsistency~\cite{Zhang2023SirensSI}; reports of model reproducibility and variability across evaluation settings (e.g., ~\cite{achiam2023gpt}).

In contrast to earlier work that viewed this variation as random noise, we show the cache-ON/cache-OFF discrepancy is systematic, reproducible, and mechanistically attributable to FP16 accumulation order, a deterministic consequence of execution path divergence rather than a stochastic phenomenon.

\section{Related Work}
Prior work has considered training of numerical precision, KV cache efficiency, mechanistic Interpretability, and LLM robustness individually.  We have not identified any prior work that quantifies execution-path difference between cache-ON and cache-OFF inference, nor investigated the interplay with FP16 accumulation order and choices like GQA ratio.  Our work relates to four bodies of prior work.

\subsection{Numerical Precision in Neural Networks}

Mixed-precision training~\cite{micikevicius2017mixed} showed that FP16 arithmetic with loss scaling can enable training speedups without accuracy loss, while~\cite{gupta2015deep} showed deep networks can tolerate limited-precision arithmetic if it comes with negligible accuracy degradation. Later work extended the application of reduced precision to inference post-training quantization~\cite{frantar2022gptq}, 8-bit matrix multiplication~\cite{dettmers2022gpt3}, activation stabilization~\cite{xiao2023smoothquant}, and activation-aware weight quantization~\cite{lin2024awq}. These works proved that neural networks are generally robust to reduced precision, albeit focusing on training stability or model compression, not on the interaction between accumulation order and KV cache execution path. No prior paper isolates the order divergence caused systematically by FP16 non-associativity on deterministic cached inference.

\subsection{KV Cache Optimization and Attention Efficiency}

Flash Attention~\cite{dao2022flashattention} and its subsequent version~\cite{dao2023flashattention} gave IO-efficient exact attention kernels that add an order of magnitude of throughput while reducing memory access for transformer decoding, both in designs implicitly requiring equality of value across different caching schemes. 
Paged Attention~\cite{kwon2023efficient} offered scalable KV cache handling at large inference scale,  explicitly optimizing for memory use without analysis of output correctness differences across execution paths. Multi-query attention~\cite{shazeer2019fast} and grouped-query attention~\cite{ainslie2023gqa} enable decoding savings by sharing keys and values across heads, but neither study considers the interactions of shared KV tensors with the FP16 accumulation order, thus overlooking impacts on output self-consistency. This is the first work that systematically measured the divergence caused by execution path variations in KV caching and demonstrated that the GQA ratio can predict the divergence magnitudes via error propagation among queries, amplified by FP16 accumulation.

\subsection{Mechanistic Interpretability}

Causal tracing and ROME~\cite{meng2022locating} introduced residual stream interventions for the discovery and editing of facts in transformers,  postulating the residual stream as the dominant site of causal intervention. Path patching~\cite{wang2022interpretability} isolated causal pathways in residual stream contributions for the indirect object identification task, while causal mediation analysis~\cite{NEURIPS2020_92650b2e} formalized activation-level interventions for mechanistic attribution. Interpretable computational primitives like induction heads were discovered over the assumption of a deterministic forward pass~\cite{olsson2022context}.  These all target the residual stream and implicitly assume that divergence stems there. Our activation patching finds a null: residual stream patching cannot recover KV cache divergence even when all layers are patched at once.  KV cache state itself emerges as the causal variable, a cause neglected by existing work.

\subsection{LLM Output Consistency and Reproducibility}

Previous work has investigated variability of LLM output due to stochastic sampling, prompt sensitivity~\cite{zheng2023judging}, dataset-specific limitations~\cite{10.1145/3442188.3445922},  or systemic reproducibility issues by comparing model versions and evaluation conditions~\cite{Chen2023HowIC, achiam2023gpt}. Those works treat source of output variability as orthogonal to the modeling as such,  depending instead solely on stochastic or distributional shift effects. Our work is fundamentally different: we elicit reproducible,  deterministic divergence from reasoning only from the execution path of the KV cache under FP16 inference with a systematic reproducibility effect that appears with greedy decoding and can be mechanistically explained by FP16 accumulation order, without the influence of any stochastic factor.

Most closely related to our work, ~\cite{yuan2025understanding} shows that FP16 and BF16 non-associativity can lead to reproducibility failures across different system configurations like batch size, GPU count, or hardware platform, and also introduces LayerCast to mitigate this, which stores weights in FP16 but 
computes in FP32. Our findings are complementary but differ in three ways: we instead show the effects of divergence between cache-ON and cache-OFF paths on \emph{identical} hardware with the same batch size, which is a deterministic effect unaffected by configuration. Secondly, we isolate the KV cache execution path exactly as the divergence source, and localize it through activation patching (which is not examined by~\cite{yuan2025understanding}). Thirdly, we analyze how the GQA ratio serves as an indicator of the divergence magnitude, which is not explored by them but is very relevant to the current LLM architectures.

\section{Analysis Framework for the Impacts of KV Cache on Divergence}\label{sec-analysis-method}

To understand KV cache divergence, knowing that there is divergence is not enough. We must understand why it occurs, how it spreads, where it is, and when it is important behaviorally. The five experiments form a causal chain in which each experiment asks a question that motivates the next.

\subsection{Experimental Logic}

\paragraph{Step 1: Does the divergence exist and is it systematic?}
We verify that cache-ON and cache-OFF produce different outputs and whether it occurs in greedy decoding, which eliminates sampling as a possible cause.  If it is not actually real and structural, then there is nothing to explain.  The measure taken in this step is the token divergence rate,  per-step KL divergence over all the model-strategy pairs.

\paragraph{Step 2: How does it propagate?}

Once divergence is confirmed, we then explore how an error in the first layer propagates through subsequent layers. This indicates if the effect is localized or evenly spread, and how the architecture's GQA ratio, head dimension, and attention type influence this propagation.

\paragraph{Step 3: What causes it?}

Propagation characterisation does not unveil the origin. We employ the FP16 non-associativity hypothesis directly: if accumulation order is indeed the cause,  converting to FP32 should eradicate divergence. The falsification discards other explanations.

\paragraph{Step 4: When does it matter behaviorally?}

Confirming the cause raises a new question: an apparent contradiction exists between the flip index and the KL magnitude. Addressing this contradiction discloses the conditions under which numerical divergence actually changes the behavior of the model, like decision boundary proximity.

\paragraph{Step 5: Where does the divergence reside?}

Finally, we localize the causal variable. If the divergence is carried out by the residual stream, patching hidden states should recover the cache-free trajectory. If patching fails, the divergence must reside in the KV cache state itself.

These five steps correspond to the five experiments: behavioral characterization (Step~1), layer drift analysis (Step~2), root cause falsification (Step~3), decision boundary analysis (Step~4), and causal localization (Step~5).

\subsection{Decoding Strategies}

The choice of decoding strategy is not incidental; it determines which alternative explanations can be ruled out. We build our own token-by-token generation loop rather than using the default with \texttt{model.generate()} API for two reasons: (1) to get full-vocab logits present at each decode step for both cache-ON and cache-OFF, so we can compute KL at each step; and (2) so that both paths have the same random seed in each step so that any differences are due to the distribution divergence, not random sampling variance, neither of them is provided in the native API.

\begin{itemize}

    \item \textbf{Greedy} ($T=0.0$): token selection is a deterministic arg max, all differences between cache-ON and cache-OFF will come solely from the upstream FP16 KV cache computation, and not the sampling, and can thus be used to isolate the pure execution-path effect.

    \item \textbf{Top-$k$ sampling} ($k=50$, $T=0.7$): Introduces a restricted stochasticity but filters off low probability words; when added to the shared seed design, this confirms the effect persists under stochastic decoding.

    \item \textbf{Nucleus (top-$p$) sampling} ($p=0.95$, $T=0.7$): Adapts the candidate set dynamically according to the confidence of the current model. This represents a quite different stochastic regime from top-$k$.

\end{itemize}

Taken together,  these three strategies show that divergence does not come from a particular decoding regime but from a fundamental property of cached inference of FP16. All three strategies were run at \texttt{max\_new\_tokens = 128} with the complete KL trajectory recorded per-step.

\subsection{Divergence Metrics}

Each metric captures a different aspect of the divergence and answers a different question in the argument.

\paragraph{Token divergence} Run pairs are divergent if generated token sequences are different at any point (binary outcome). This directly establishes whether cache-ON and cache-OFF generate different text. It is the primary behavioral outcome of Step~1: if token divergence is universal, the effect is structural rather than incidental.


\paragraph{KL divergence} At each decoding step, the KL divergence is calculated between cache-ON distribution $p_t$ and cache-OFF distribution $q_t$ over the whole vocabulary:

Let $p_t$ and $q_t$ denote the softmax distributions from cache-ON and cache-OFF at step $t$, respectively.
\begin{equation}
    \text{KL}(p_t \| q_t) = \sum_{v} 
    p_t(v) \log \frac{p_t(v)}{q_t(v)}
\end{equation}
Mean KL across steps is the primary magnitude measure. KL measures the amount of probability mass moved between the two paths, even before the token is chosen. A consistently high KL trajectory indicates that the distributions diverge continuously, not merely fluctuate. 
This enables layer-by-layer drift analysis in Step~2 and the FP32 magnitude comparison in Step~3.



\paragraph{JS divergence} JS divergence is the symmetric Jensen-Shannon divergence between $p_t$ and $q_t$:


\begin{equation}
    \text{JS}(p_t \| q_t) = \frac{1}{2}
    \text{KL}(p_t \| m_t) + \frac{1}{2}
    \text{KL}(q_t \| m_t)
\end{equation}
where $m_t = \frac{1}{2}(p_t + q_t)$,bounded in $[0, 1]$.

If JS reflects the same trend as KL, then the divergence pattern is presumably valid, rather than a mathematical illusion coming from KL‘s unbounded range. JS serves as a consistency check on the KL findings.

\paragraph{Flip index} The flip index is the first decode step at which cache-ON and cache-OFF differentiate on a token.  This is the point where the behavioral divergence begins, when numerical drift first becomes observable in generated text. Its relationship with mean KL is explored in Step~4: an apparent contradiction between early flips and high KL reveals that decision boundary proximity, not divergence magnitude, governs when divergence becomes important.


\subsection{Statistical Analysis}
Each test is selected to answer a particular inferential question in the argument rather than as a more general check of significance.

\paragraph{Behavioral significance}

Applies McNemar‘s test to the pairwise correctness results (cache-ON correct vs.\ cache-OFF correct on the same example).  Pairing by example controls for prompt difficulty; the test asks specifically whether one execution path systematically outperforms the other, not merely whether accuracy differs. This is a secondary criterion for Step~1 as we need to verify that the direction of divergence is consistent, not random. Multiple comparisons are corrected using Bonferroni and BH-FDR at $\alpha = 0.05$; bootstrap 95\% confidence intervals are given for mean KL and delta in accuracy.

\paragraph{Layer drift significance} runs Mann--Whitney U tests on per-layer magnitudes of hidden state drift over the 90-layer-model combinations, with BH-FDR correction. The non-parametric test is chosen because drift magnitudes are right-skewed and not normally distributed. This aids Step~2 by indicating which layers then statistically reliably amplify divergence.

\paragraph{Step-count correlation} uses Pearson and Spearman correlations between flip index and mean KL divergence ($n = 10{,}500$ per model). Since the flip index distribution is heavily skewed toward zero, both correlations are listed; Spearman rank correlations are also presented as a robustness check. Partial correlations excluding flip index $= 1$ and $\leq 3$, test whether the negative correlation is stronger without the most extreme boundary flips, providing a direct test of step~4.

\section{Experiments and Data Collection}\label{sec-experiments}

\subsection{Experiment Environment}

\paragraph{Models}
We study three open-weight autoregressive models of different architectural classes: LLaMA-2-7B ~\cite{touvron2023llama}, Mistral-7B-v0.3 ~\cite{Jiang2023Mistral7}, and Gemma-2-2B ~\cite{team2024gemma}. We chose these models to sample across the axes along which major architectural differences can be characterized: attention type (GQA versus MHA), GQA sharing ratio (1:1 for LLaMA-2, 4:1 for Mistral, 2:1 for Gemma), head dimension (128 versus 256), sliding window attention (Gemma only), and parameter count (2B versus 7B) to find systematic effects. We load all models in FP16 using HuggingFace \texttt{AutoModelForCausalLM.from\_pretrained} unless specified otherwise.


\paragraph{Datasets}
We evaluate on the GSM8K mathematical reasoning benchmark~\cite{cobbe2021training}, applying measurements to 700 randomly sampled examples without replacement (\texttt{SAMPLE\_SEED=42}) from the provided test set of 1,319. This benchmark was used because it contains multi-step mathematical problems that are sufficiently long to involve trajectories with multiple KV cache compositions and provides unambiguous binary correctness criteria. Each test example was evaluated under five different sample seeds (0-4) for a total of 3,500 runs per model-strategy combination, and 31,500 overall across all nine combinations of model and decoding strategy.  We have 16 pairwise comparisons with achieved power $\geq 0.97$ at $n = 3,500$; the two underpowered comparisons (LLaMA-2 greedy vs top-$p$ and Gemma top-$k$ vs top-$p$) are treated as exploratory and excluded from confirmatory claims. 

We note that the absolute accuracy numbers are rather low given our protocol: \texttt{max\_new\_tokens=128} may truncate some generations before they can be completed, and our few-shot prompt format can cause models to mimic in-context exemplars. The accuracy difference between cache-ON and cache-OFF should therefore be interpreted as an indicator of execution-path bias, not a measure of absolute reasoning ability.  Our primary behavioral finding is that cache-ON and cache-OFF differ in 100\% of cases, and that the direction is systematic rather than random, and does not rely on absolute accuracy levels.
\paragraph{System environment}

All experiments are conducted on a single NVIDIA H100 NVL (95 GB) GPU, paired with an AMD EPYC 9254 24-core CPU and 250 GB RAM. The system runs Ubuntu 22.04 with Python 3.12.3, PyTorch 2.5.1 (CUDA 12.1), and HuggingFace Transformers 4.53.0.

For greedy decoding, we set
\texttt{CUBLAS\_WORKSPACE\_CONFIG=:4096:8}
and enable deterministic behavior via
\texttt{torch.use\_deterministic\_algorithms(True)},
ensuring bitwise reproducibility across reruns on the same GPU.

\subsection{Experiment Design}

The five experiments follow the causal logic established in the analysis framework (Section~\ref{sec-analysis-method}). Each is described below with its role in the argument.

\paragraph{Behavioral Characterization.} The first question is whether divergence exists and is structural. For each of the 31,500 pairs of runs, we record the token divergence rate, cache-ON and cache-OFF mean KL divergence, flip index, and others. The reproducibility of each example is evaluated by whether it always-diverges, sometimes-diverges, or never-diverges across five seeds, establishing whether divergence is a property of the example or the seed.

\paragraph{Layer Drift Analysis.}Having confirmed divergence is universal, we ask how it propagates through the network. For a subset of high-divergence examples, hidden state representations for each transformer layer are captured under both cache-ON and cache-OFF during generation. Per-layer drift is reported at decode step~1 for the primary layer analysis and computed via the L2 norm of the hidden state difference vector and cosine similarity between cache-ON and cache-OFF states. Per-head attention weight KL divergence is also isolated to verify whether divergence appears in attention patterns or in accumulated hidden values.


\paragraph{Root Cause Falsification.} Layer drift characterizes propagation but does not identify the cause. Five sub-experiments test the FP16 non-associativity hypothesis. Experiment 1 quantifies the relative FP16 accumulation error of dot-products vs.\ sequence length ($\approx 0.036\%$, flat across all lengths and models, std $= 8.5 \times 10^{-7}$). Experiment 2 simulates multi-layer error propagation through $N$ sequential attention and residual steps, showing how single-layer FP16 rounding errors compound across layers, consistent with the large KL divergence observed in the behavioral characterization. Experiment 3 measures the FP16 rounding error in $K$ and $V$ projections between joint computation (cache-OFF) and token-by-token computation (cache-ON) to confirm the error originates at cache write time rather than downstream. Experiment 4 correlates the flip index with the mean KL divergence to test whether decision boundary proximity mediates the effect. Experiment 5 reruns the full inference pipeline in FP32 for $n=600$ examples per model; if FP16 accumulation order is the cause, FP32 must eliminate divergence. Flip rate is $0.0$ across all models, providing a clean falsification.

\paragraph{ Decision Boundary Analysis.} Having confirmed the root cause, we address a seeming contradiction: examples that flip earlier have greater KL divergence. Seems to go against the accumulation hypothesis. Run pairs are split into early-flip (bottom 33rd percentile of flip index) and late-flip (top 33rd percentile) groups; Welch's $t$-test confirms that early-flip examples have \emph{higher} mean KL in 8 of 9 conditions. Pearson correlation between flip index and mean KL is consistently negative across models ($r = -0.202$ to $-0.120$, all $p < 10^{-34}$). Partial correlations excluding flip index $= 1$ and flip index $\leq 3$ test whether the negative correlation strengthens after removing boundary cases.

\textbf{Activation Patching.} The final question is where the divergence resides. For the $n = 600$ highest-KL examples per model, (selected in decreasing order of mean KL from the full 700), cache-ON hidden states are injected into the cache-OFF forward pass at decode step 0, as it is the first step at which the hidden states diverge, as the prefill produces the identical states under both paths ($\|\mathbf{h}_{\text{on}} - \mathbf{h}_{\text{off}}\| = 0$ at all layers).
This is done one layer at a time and cumulatively across all layers $0$ to $L$. Recovery is measured as:
\begin{equation}
    \text{pct\_recovered} = 
    \frac{\text{KL}_{\text{base}} - 
    \text{KL}_{\text{patched}}}
    {\text{KL}_{\text{base}}} \times 100
\end{equation}
Negative recovery here confirms that patching increases divergence. All patching experiments are run with \texttt{MAX\_STEPS=32} decode steps per example.

\section{Experiment Results Analysis}\label{sec-data-analysis}

We present findings in five parts following the causal logic of the experimental design.

\subsection{Behavioral Characterization}

Table~\ref{tab:q3} summarizes divergence rate, 
mean KL divergence, and task accuracy across 
all nine model-strategy conditions.

\begin{table}[ht]
\centering
\caption{Behavioral results across all nine 
model--strategy conditions ($n = 3{,}500$ run 
pairs per condition). Divergence rate is 100\% 
in every condition without exception. Mean KL 
and 95\% CI are bootstrap estimates over run 
pairs. Cache acc.\ and No-cache acc.\ are 
indicators of execution-path bias under our evaluation 
protocol, not measures of absolute reasoning ability. $\dagger$ denotes 
Bonferroni-significant McNemar result 
($p < 0.05$); $*$ denotes non-significant 
(exploratory).}
\label{tab:q3}
\resizebox{\columnwidth}{!}{%
\begin{tabular}{llccccc}
\toprule
Model & Strategy & Div.\ Rate & Mean KL 
& KL 95\% CI & Cache Acc. & No-Cache Acc. \\
\midrule
LLaMA-2 & Greedy  & 100\% & 10.68 
& [10.63, 10.74] & 3.3\% & 2.0\%$^\dagger$ \\
LLaMA-2 & Top-$k$ & 100\% & 10.70 
& [10.65, 10.74] & 3.3\% & 1.2\%$^\dagger$ \\
LLaMA-2 & Top-$p$ & 100\% & 10.79 
& [10.75, 10.83] & 2.8\% & 1.7\%$^\dagger$ \\
\midrule
Mistral  & Greedy  & 100\% & 11.46 
& [11.40, 11.53] & 6.6\% & 2.0\%$^\dagger$ \\
Mistral  & Top-$k$ & 100\% & 10.70 
& [10.66, 10.76] & 6.5\% & 1.7\%$^\dagger$ \\
Mistral  & Top-$p$ & 100\% & 10.90 
& [10.85, 10.94] & 5.0\% & 1.9\%$^\dagger$ \\
\midrule
Gemma    & Greedy  & 100\% & 12.66 
& [12.62, 12.70] & 1.9\% & 2.1\%$^*$ \\
Gemma    & Top-$k$ & 100\% & 12.80 
& [12.76, 12.84] & 3.9\% & 1.1\%$^\dagger$ \\
Gemma    & Top-$p$ & 100\% & 12.83 
& [12.79, 12.86] & 3.3\% & 1.3\%$^\dagger$ \\
\bottomrule
\end{tabular}}
\end{table}

\textbf{Token divergence is universal.} cache-ON and cache-OFF generate distinct sequences for all 31,500 pairs across every single run (i.e., 100\% divergence rate, without a single exception). Importantly, this is under greedy decoding, which freezes no sampling at all: the next token is always the argmax of the output distribution,  so the problem can‘t be credited to stochasticity. The 100\% rate of divergence under greedy shows that the effect is a structural property of the FP16 cached inference.

\textbf{Execution-path bias is systematic, 
not random.} In 8 of the 9 conditions, cache-ON returns more correct answers than cache-OFF, and the differences reach Bonferroni corrected significance (McNemar test, $p<0.05$).  Only Gemma greedy has non-significant differences ($p=0.43$) and the direction is inverted, cache-OFF edges cache-ON (2.1\% vs 1.9\%), suggesting that the greedy execution path for Gemma does not reliably give one mode an accuracy advantage. The accuracy difference is an indicator of execution-path bias under our evaluation protocol, not a measure of absolute reasoning ability. The reliable trend in 8 of 9 conditions is what is interesting, not the absolute accuracy scores.

\textbf{KL divergence is large, stable, and 
model-specific.}  Mean KL varies from 10.68 (LLaMA-2 greedy) to 12.83 (Gemma top-$p$),  yet narrow bootstrap 95\% CIs verify these values are stable across the 3{,}500 run pairs per condition.  Nonetheless,  Gemma always exhibits the most divergence across strategies. In contrast, within-model KL variation across strategies is small for LLaMA-2 and Gemma,  but not for Mistral (greedy: 11.46,  versus top-$k$ 10.70), consistent with Mistral‘s 4:1 GQA ratio, which amplifies per-step accumulation errors more aggressively than LLaMA-2's MHA structure. Mean KL $> 10$ across all conditions indicates substantial redistribution of probability mass at every decode step; these are not small numerical perturbations. JS divergence value,  which is provided as a bounded stability test ($\text{JS} \in [0,1]$),  gives the same trend as KL divergence in all conditions,  confirming the divergence pattern is not an artifact of KL's unbounded range.

Figure~\ref{fig:kl_traj} shows the KL divergence trajectory across all 128 decode steps for all three models and all three strategies. The KL divergence remains consistently high across all 128 steps without converging to zero, as expected due to the FP16 rounding errors accumulating at each decode step rather than cancelling out.

\begin{figure}[h]
\centering
\includegraphics[width=\columnwidth]{Images/Q1/kl_trajectory_all_models.png}
\caption{Mean KL divergence between cache-ON 
and cache-OFF output distributions at each 
decode step across all models and strategies. 
Divergence persists and accumulates across 
the full 128-step trajectory with no 
convergence toward zero. Gemma shows 
systematically higher divergence than LLaMA-2 
and Mistral, consistent with its GQA 
architecture and a distinct head dimension.}
\label{fig:kl_traj}
\end{figure}

\subsection{Layer Drift Analysis}
\begin{figure}[h]
\centering
\includegraphics[width=\columnwidth]{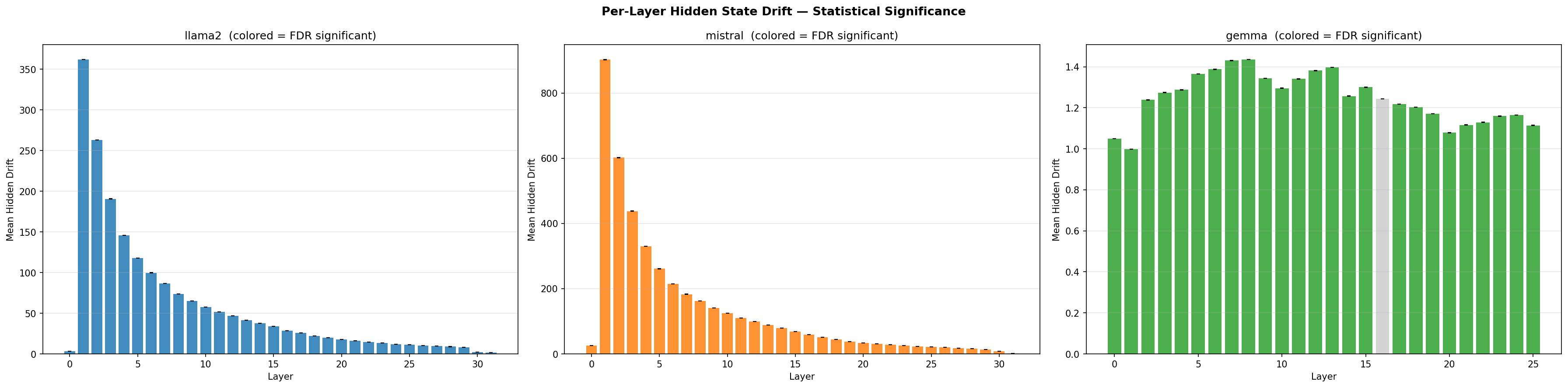}
\caption{Per-layer hidden state drift magnitude 
(L2 norm of the difference vector between 
cache-ON and cache-OFF hidden states) at decode 
step 1. LLaMA-2 and Mistral show sharp 
amplification after the first transformer 
layer. Gemma shows a qualitatively distinct 
flat profile with no amplification event. 
All 89 of 90 layer-model combinations are 
statistically significant (MWU test, BH-FDR 
corrected, $\alpha = 0.05$); only Gemma 
layer 16 is not ($p = 0.877$).}
\label{fig:layer_stats}
\end{figure}

Having established that divergence is universal and systematic, we now ask how it propagates through the network. The position dimension is well represented by the KL trajectory as shown in Figure~\ref{fig:kl_traj},  which shows divergence persisting across all decode steps; Figure~\ref{fig:layer_stats} characterizes the layer dimension.


\textbf{LLaMA-2 and Mistral show sharp 
divergence amplification at layer 1.} The LLaMA-2 hidden state drift jumps from $3.45$ for layer 0, to $361.96$, for layer 1 (which is a $105 \times$ amplification for a single layer).  Mistral even further leaps from $25.76$ for layer 0, to $902.36$ for layer 1 ($35 \times$). This gain is consistent with Mistral‘s 4:1 GQA ratio: each shared KV head spreads its FP16 rounding error over four query heads simultaneously; we get correlated errors rather than independent ones, and the initial perturbation is amplified with greater intensity than the MHA inside of LLaMA-2.

\textbf{Gemma exhibits a structurally distinct 
flat profile.} In contrast to LLaMA-2 and Mistral, there is no sudden amplification event in Gemma, but instead the drift is between 1.0 and 1.4 throughout all 26 layers. This is qualitatively different from LLaMA-2 or Mistral‘s divergence propagation pattern. Even though Gemma has the lowest per-layer drift values, it has the largest total KL divergence in the behavioral characterization, which suggests that Gemma‘s divergence is distributed across all the layers, rather than being centralized to one point of amplification. 25 out of 26 layers are significant for Gemma (BH-FDR), except for layer 16 (p = 0.877). The flat profile Gemma exhibits is consistent with the larger head dimension (256 compared to 128) and the sliding window attention in that it also only allows each layer to see local context and not the full sequence (with the associated limitation on allowing errors to propagate globally), hence the slow build-up of divergence across all layers rather than any amplification at a specific point.

\textbf{Cosine similarity confirms rapid 
decorrelation.} This cosine similarity of cache-ON and cache-OFF hidden states falls to 0.49 in layer 0 for LLaMA-2 in decode step 1 and falls to 0.015 for Mistral, near-complete decorrelation in the first transformer layer already, aligning with Mistral‘s large amplification of layer-1 drift. Gemma starts from a higher value of 0.85, going down to nearly 0 by layer 13 (cosine = $0.009$), oscillating non-monotonically on middle layers, and recovery up to some extent in later layers, aligning with its gradual flat drift profile. These variations in propagating divergence reflect different architectural propagations in attention-based models, directly impacting the causal localization analysis.

\subsection{Root Cause Falsification}

Layer drift analysis shows how FP16 rounding errors propagate but does not identify their origin. We now test whether FP16 accumulation order is the cause via direct falsification: if it is, running the same pipeline in FP32 should eliminate divergence.

\begin{figure}[h]
\centering
\includegraphics[width=\columnwidth]{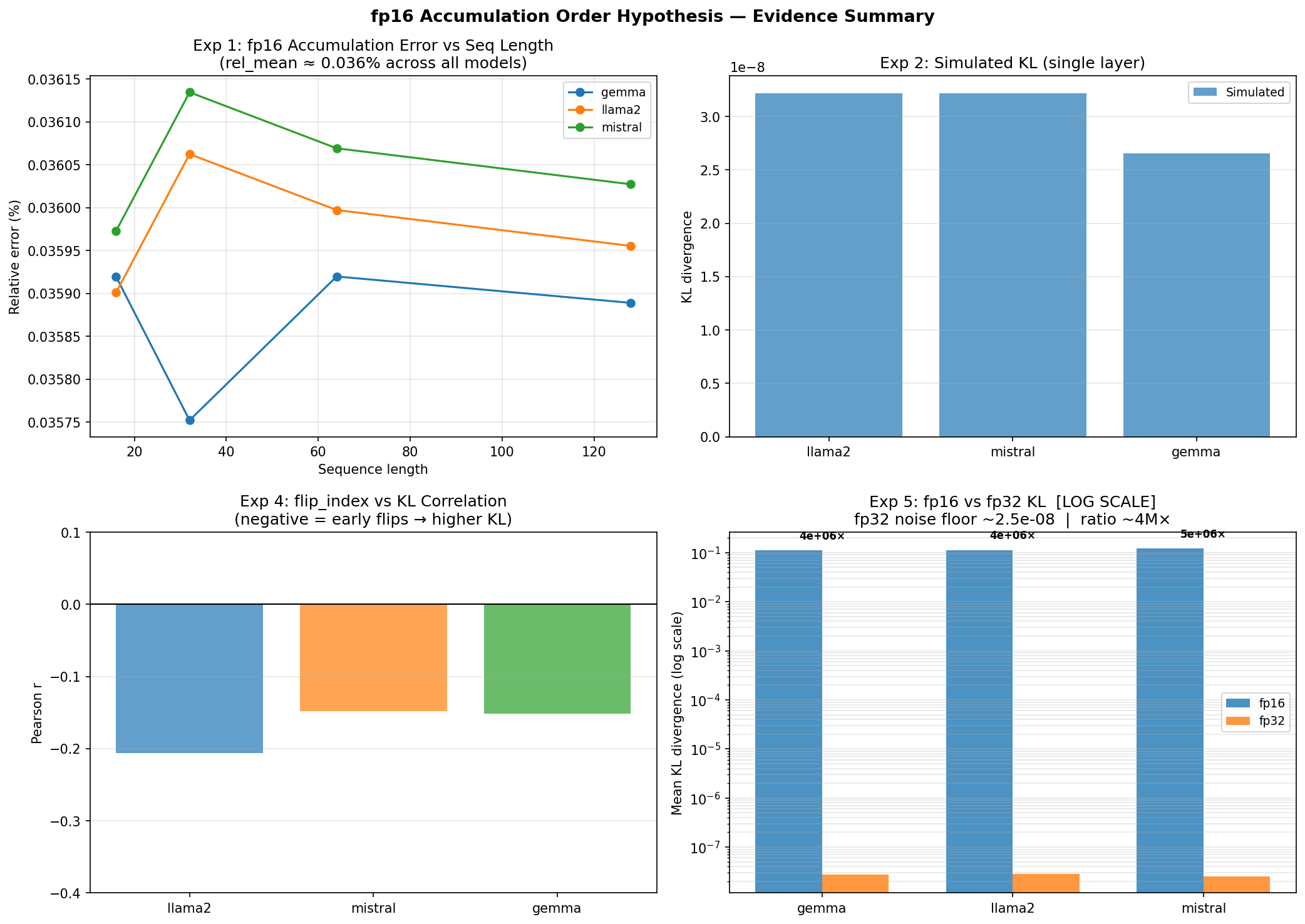}
\caption{FP16 versus FP32 KL divergence 
across all three models ($n = 600$ examples 
per model). Under FP32, KL divergence 
collapses to a noise floor of 
${\sim}2.5 \times 10^{-8}$ and flip rate 
drops to exactly 0.0, providing clean 
falsification of the FP16 accumulation 
order hypothesis. Mean relative accumulation 
error is flat across all sequence lengths 
($\approx 0.036\%$), confirming it is 
architecture-determined.}
\label{fig:hypothesis}
\end{figure}

\begin{table}[ht]
\centering
\caption{FP16 versus FP32 KL divergence and 
flip rate per model ($n = 600$ examples, 
32 decode steps). FP16 mean KL is computed 
over all examples. Under FP32, flip rate 
is exactly 0.0 for every example across 
all three models.}
\label{tab:fp32}
\begin{tabular}{lcccc}
\toprule
Model & FP16 KL & FP32 KL & FP16 Flip 
& FP32 Flip \\
\midrule
LLaMA-2 & $0.113$ & $2.80 \times 10^{-8}$ 
& $1.15\%$ & $0.0\%$ \\
Mistral  & $0.122$ & $2.50 \times 10^{-8}$ 
& $1.14\%$ & $0.0\%$ \\
Gemma    & $0.112$ & $2.78 \times 10^{-8}$ 
& $1.03\%$ & $0.0\%$ \\
\bottomrule
\end{tabular}
\end{table}

\textbf{FP32 eliminates divergence entirely.} Under FP32 KL divergence drops to a noise floor around $2.5 \times 10^{-8}$ in all three models (Figure~\ref{fig:hypothesis}, Table~\ref{tab:fp32}). Flip rate is precisely 0.0 for all examples in all 3 models: no token sequence at all is altered between cache-ON and cache-OFF under FP32. This is a clean falsification: divergence does not lie in the mechanism of the KV cache itself, but in the format (FP16) chosen to implement it.

\textbf{The reduction spans more than eight 
orders of magnitude.} The accumulated FP16 full-model KL averages ${\sim}10^{1}$ across conditions. The FP32 noise floor is in the vicinity of ${\sim}2.5 \times 10^{-8}$.  This ratio of ${\sim}4 \times 10^{8}$ exceeds eight orders of magnitude; there is no ambiguity here:  FP16 non-associativity is the dominant and sufficient cause of the observed divergence.

\textbf{Accumulation error is 
architecture-determined.} The average relative FP16 accumulation error is $ 0.036 \%$, it remains relatively flat for all input sequence lengths (16-128 tokens), and it is identical for all three models. This implies the error norm is determined by the attention head dimension rather than input data or length, which means the divergence can be predicted from architecture alone without knowledge of the input.

\subsection{Decision Boundary Analysis}

\begin{figure}[h]
\centering
\includegraphics[width=\columnwidth]{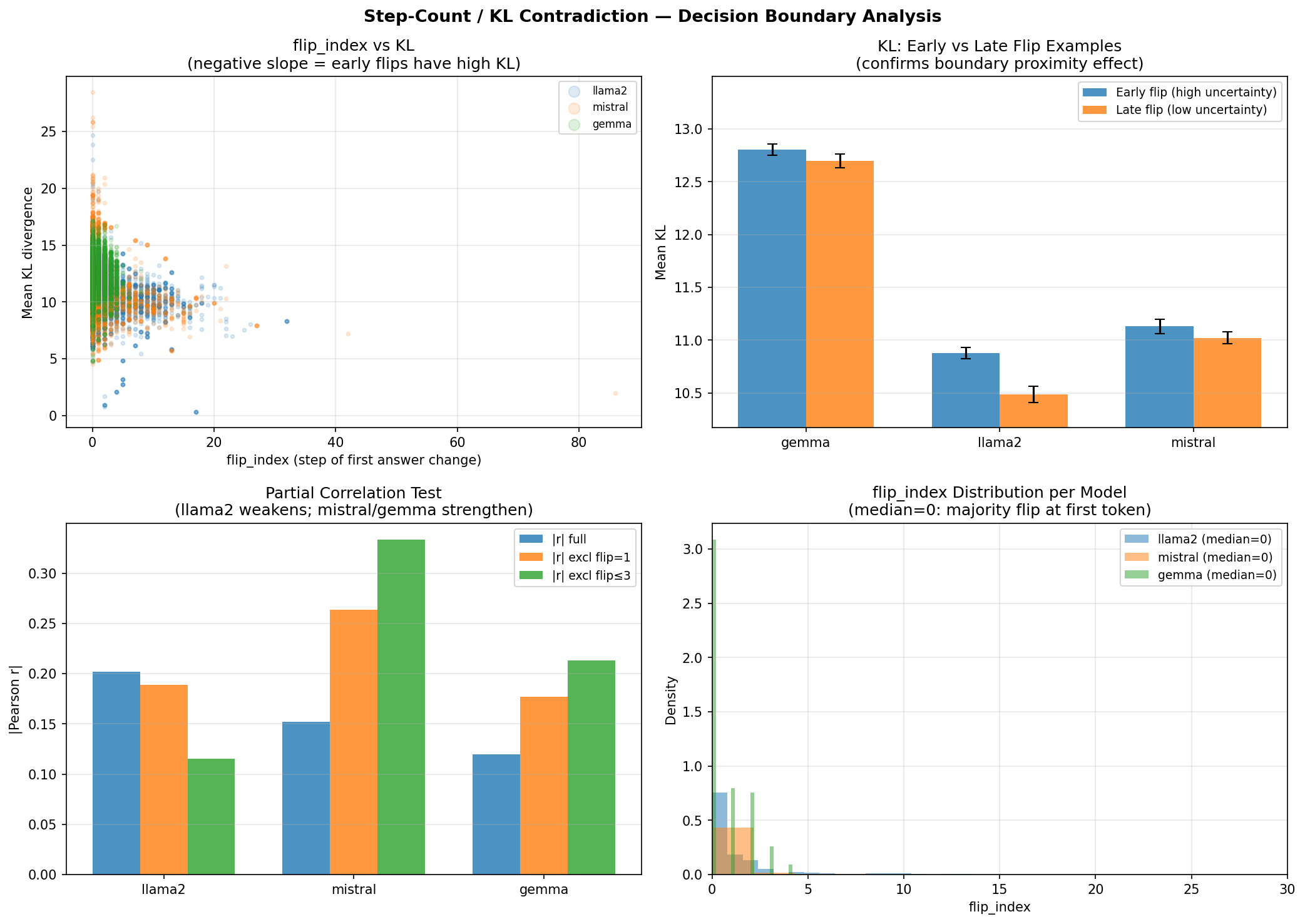}
\caption{Decision boundary analysis: four views of the flip index--KL relationship. \textbf{Top-left:} Scatter of flip index versus mean KL divergence; the negative slope confirms that examples flipping earlier have higher KL. \textbf{Top-right:} Mean KL for early-flip (high uncertainty) versus late-flip (low uncertainty) examples per model; early-flip KL is consistently higher, confirming decision boundary proximity as the explanation. \textbf{Bottom-left:} Absolute Pearson $|r|$ under three exclusion conditions; LLaMA-2 weakens after removing boundary cases while Mistral and Gemma strengthen. \textbf{Bottom-right:} Distribution of flip index per model; median $= 0$ across all three models confirms the majority of flips occur at the first generated token.}
\label{fig:stepcount}
\end{figure}

Root cause falsification confirms FP16 non-associativity as the cause, but an apparent contradiction remains: examples that flip earlier have higher KL. Here, Figure~\ref{fig:stepcount} presents four complementary views of this relationship and its resolution; Table~\ref{tab:p3} reports the Pearson correlations between the flip index and the KL divergence across all models and strategies.


\begin{table}[ht]
\centering
\caption{Pearson correlation between flip 
index and mean KL divergence ($n = 10{,}500$ 
per model), with partial correlations 
excluding early-flip boundary cases. All 
full correlations are significant 
($p < 10^{-34}$). Mistral and Gemma show 
a strengthening correlation after removing 
boundary cases, confirming the decision 
boundary proximity hypothesis.}
\label{tab:p3}
\begin{tabular}{lcccc}
\toprule
Model & Full $r$ & Excl.\ flip $=1$ 
& Excl.\ flip $\leq 3$ & Trend \\
\midrule
LLaMA-2 & $-0.202$ & $-0.189$ & $-0.115$ 
& weakens \\
Mistral  & $-0.152$ & $-0.264$ & $-0.333$ 
& strengthens \\
Gemma    & $-0.120$ & $-0.177$ & $-0.213$ 
& strengthens \\
\bottomrule
\end{tabular}
\end{table}
\textbf{That early flips associate with higher KL is an apparent contradiction.}
Figure~\ref{fig:stepcount} shows that Pearson correlation is negative for all 3 models with flip index and mean KL. $r = -0.202$ (LLaMA-2, $p < 10^{-96}$), $r = -0.152$ (Mistral, $p < 10^{-55}$), $r = -0.120$ (Gemma, $p < 10^{-34}$). Examples that flip earlier have a higher KL divergence. This appears to contradict the accumulation hypothesis, which predicts that more accumulated divergence should cause earlier flips.

\textbf{Decision boundary proximity resolves 
the contradiction.} KL for early-flip examples is higher than for late-flip examples for 8 out of 9 conditions ($t$-test, $p<0.05$), with ratios in the range $(1.006-1.043)$.  The reason behind this is the distance to the decision boundary: examples close to the decision boundary will have small logit margins,  and flip on the first decoding step, so that any perturbation crosses the threshold.  These same examples have high KL as a consequence of the uncertainty spreading mass out over the tokens.  Early flips and high KL are joint consequences of uncertainty, not causally linked to each other.

\textbf{The correlation strengthens after 
removing boundary cases for GQA models.} Partial correlation (without flip index $\leq 3$) strengthens negative correlation for Mistral ($-0.152 \rightarrow -0.333$) and Gemma ($-0.120 \rightarrow -0.213$),  ruling out outliers for boundary case driving effect (Table~\ref{tab:p3}). LLaMA-2 weakens ($-0.202 \rightarrow -0.115$), consistent with differences in the architecture between the MHA and GQA model in how boundary uncertainty propagates across decode steps.

\subsection{Causal Localization}

Having confirmed the cause and characterized propagation, the final question is where the divergence resides. If it lives in the residual stream, patching hidden states should recover the cache-free trajectory.

\begin{figure}[h]
\centering
\includegraphics[width=\columnwidth]{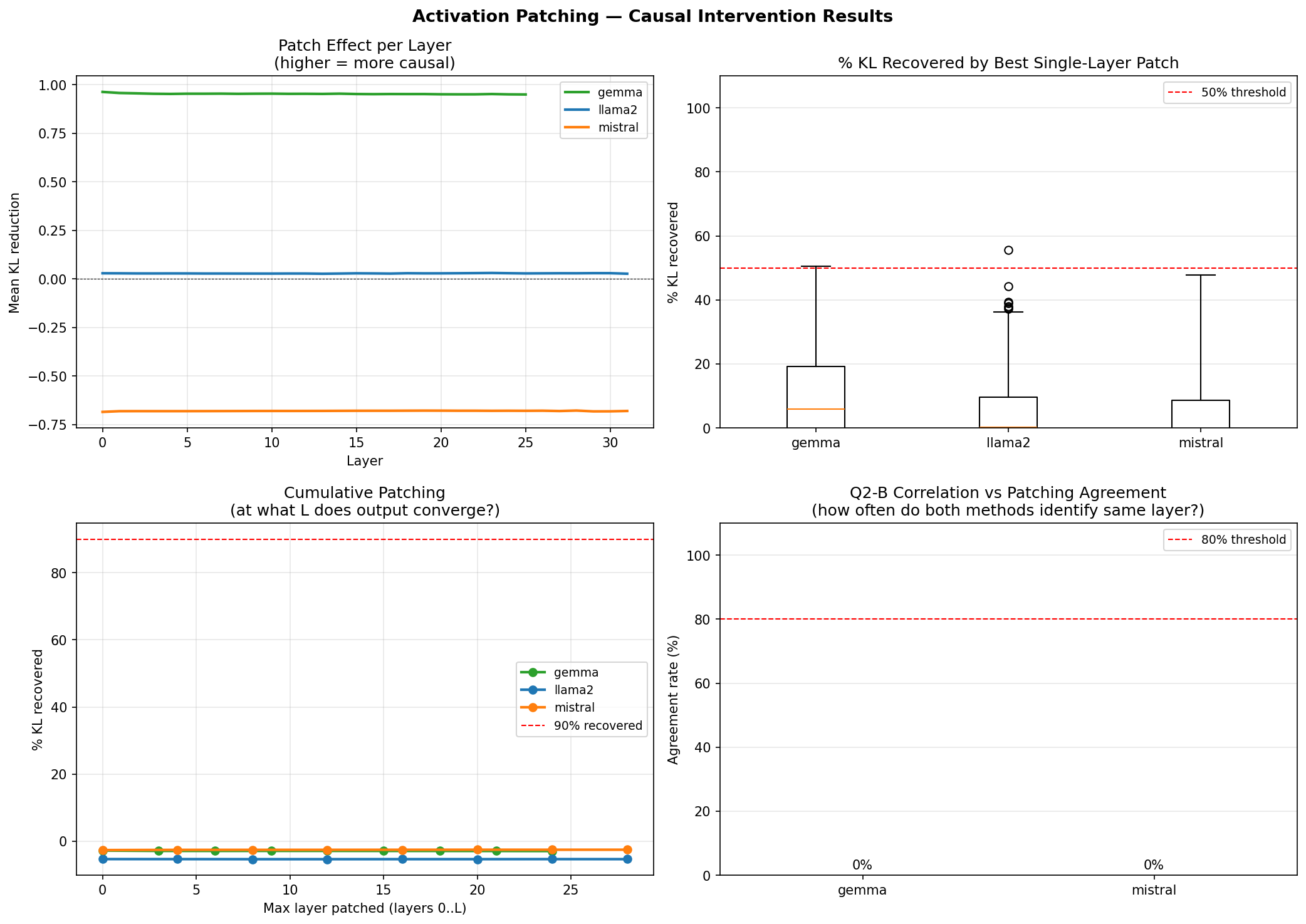}
\caption{Activation patching results for all 
three models. Each point shows the percentage KL 
divergence recovered by patching a single 
layer's cache-ON hidden state into the 
cache-OFF forward pass at decode step 0. 
Overall median recovery is $-0.019\%$ across 
all layers and models. Cumulative patching 
of all layers simultaneously produces 
negative mean recovery in all three models, 
localizing the causal variable to the KV 
cache state rather than the residual stream.}
\label{fig:patching}
\end{figure}

\textbf{Residual stream patching cannot recover KV cache divergence.} 
As shown in Figure~\ref{fig:patching}, single-layer patching results in a net median recovery of $-0.019\%$ across all layers and models (zero on average),  with a mean of $-1.21\%$,  showing that patching does not on the whole result in convergence to the target divergence. The high variance of individual layer recoveries (LLaMA-2: -107\% to + 56\%) suggests that single-layer patches are perturbing rather than correcting the divergence trajectory. All three models have a negative mean recovery when patching all layers simultaneously,  LLaMA-2: $-5.34\%$,  Mistral: $-2.56\%$,  Gemma: $-2.88\%$, when patching the entire residual stream.

\textbf{Gemma shows partial single-layer recovery, consistent with its flat layer drift profile.} 
Gemma‘s one-layer median recovery is +5.75\%, compared to LLaMA-2 (+0.08\%) and Mistral (-8.83\%).  This is an actual architectural difference, not just noise: across all 26 Gemma layers, this positive recovery holds steady (median range is $5.73\%$--$5.85\%$) and follows directly from the flat layer drift profile of Gemma. Because Gemma‘s divergence accumulates steadily and uniformly across all layers and does not spike abruptly at layer 1, single-layer patches can partially redirect the local divergence trajectory.  In LLaMA-2 and Mistral, the layer-1 acceleration ($105\times$ and $35\times$ respectively) means that a single patch intervention can have no discernible impact locally.

\textbf{Cumulative patching confirms causal localization for all three models.} 

However, the cumulative patching of all layers simultaneously still results in a negative mean recovery ($-2.88\%$) for Gemma as well. This is the final result: simultaneously patching all layers of the residual stream is not sufficient to repair the divergence. These results are consistent with the causal variable being in the KV cache state: the FP16 rounding error written to KV tensors on each decode step stays in cached KV states and affects all downstream attention computation, but residual stream interventions have no visibility to it. Correcting the divergence requires direct patching of the KV cache tensors, not the hidden states.

\section{Discussion}

\subsection{A Unified Mechanistic Account}

The five experiments form one unifying argument. Behavioral characterization shows cache-ON and cache-OFF always diverge in FP16 in every run, model, strategy, and seed. Root cause falsification identifies the cause: FP16 non-associativity causes two execution paths to accumulate in a different order, the divergence being eliminated in FP32. Layer drift analysis demonstrates how the initial error gets passed around. The error first grows rapidly at layer 1, in both LLaMA-2 and Mistral, and is broadly distributed across all layers in Gemma.  Decision boundary analysis shows when this matters behaviorally: not when the divergence is large in magnitude, but when the model is in the vicinity of a decision boundary. Causal localization establishes exactly what component the error lives in: patching the residual stream entirely fails, even when all layers are patched at once, indicating that the KV cache state is carrying the error forward.

This mechanism is deterministic, reproducible, architecturally predictable, and non-correctable by post-processing, which clearly differs from the stochastic fluctuations in output demonstrated in prior studies~\cite{zheng2023judging, Chen2023HowIC}, which depend on sampling temperature and prompt sensitivity, and the batch-size and hardware-dependent nondeterminism recently characterized by~\cite{yuan2025understanding}. Our divergence requires neither: it appears on identical hardware with identical batch size; this phenomenon exists even under greedy decoding with a fixed seed, and can be eliminated by changing the arithmetic precision.


\subsection{Architectural Predictability}

Divergence is not a random phenomenon; it could be predicted from the architecture of attention. The $0.036\%$ average relative accumulation error is flat over sequence length and models, establishing that per-step error magnitude is head dimension and attention structure-determined, not input context. When combined with GQA, this becomes significantly worse, as one FP16 error propagates to $R$ query heads and therefore $R$ is correlated instead of a single independent error term. For within-model layer-1 amplification, this yields $105{\times} $ for LLaMA-2 and $35\times$ for Mistral (since Mistral‘s training maintained larger baseline drift, the absolute divergence at layer 1 is actually larger, 902 versus 362, and it is still consistent with the idea that GQA broadcasts error across four query heads simultaneously.) Gemma‘s 256-dimensional heads have a clear flat propagation pattern and broad uniform accumulation rather than focused amplification that leads to the highest overall KL divergence despite having the lowest per-layer drift. This all depends solely on architecture and should be predictable before inference.

\subsection{The Localization Argument and Its Limits}

Causal localization establishes that divergence does not reside in the residual stream since we find the residual stream can‘t recover it, even with the entire set of layers patched. The obvious candidate is the KV cache state itself, where at each decode step, FP16 rounding errors are written and carried forward. This is an elimination argument: a proof by direct patching of the KV cache tensor would be better. We have identified this as the most important immediate extension. The current results establish a necessary condition: any residual stream intervention will be unable to recover the divergence.

\subsection{Implications}

\paragraph{Cache equivalence cannot be assumed.}
All FP16 production systems leveraging KV caching yield distinct results from no-cache inference -- in $100\%$ of the cases. The validation pipelines checking cached and non-cached results will report these differences, but as properties intrinsic to FP16, and not errors.

\paragraph{Divergence direction is systematic.} 
Under our evaluation protocol, cache-ON yields higher accuracy than cache-OFF in 8 of 9 conditions, here this accuracy difference indicates a systematic execution-path bias rather than random divergence.
Task systems that use prompt caching or KV reuse can expect the outputs from the cache prefixes to systematically differ from the ones recomputed,  and the direction will depend on the architecture.

\paragraph{GQA amplifies the effect.}
Such divergence will be even more common with the advent of GQA and MQA, which are increasingly dominant as modern LLMs become larger.  Architects should consider divergence amplification in addition to savings in memory bandwidth when selecting an attention variant.

\paragraph{FP32 is the only current fix.} 
For pipelines where cache-ON/cache-OFF equivalence is necessary, FP32 reduces the divergence to a noise floor more than eight orders of magnitude smaller than FP16. It is currently one of the options where equivalence is more significant than performance.

\subsection{Limitations}

\paragraph{Scope.} 
Three models and one benchmark present important architectural dimensions, though not the full spectrum of those in use.  Naturally, larger models, instruction-tuned models, and other tasks would extend this. The divergence effect is a property of floating point arithmetic, not of the task or model, and we expect it to be true for any FP16 inference system that makes use of KV caching.


\paragraph{Evaluation protocol.} 
In the current protocol, the absolute accuracy values are low due to the truncation with \texttt{max\_new\_tokens=128} and the format of the prompt for a few-shot conditions. In addition, accuracy differences serve as indicators of execution-path bias in these conditions, not as measures of reasoning performance. The core behavioral results of all conditions divergent (100\%) and the direction systematic in 8 of 9 conditions, do not rely on the level of absolute accuracy.

\paragraph{Localization by elimination.} 
Causal localization demonstrates that divergence does not reside in the residual stream, but it does not directly prove that it resides in the KV cache. Direct patching of the KV cache is identified as future work.

\subsection{Future Work}

The most proximal extension is direct KV cache tensor patching:  at each decode step, swap cache-OFF KV tensors for cache-ON tensors and detect whether the deviation is recovered.  It turns the elimination argument from causal localization into quantitative causal evidence and is the single experiment most wanted to finalize the mechanistic story.

Another natural extension is to directly track the cache runtime ourselves, and measure how FP16 rounding errors build up over time in the K and V tensors over the course of decoding steps and layers.  That would afford a direct sense of how much errors grow in cache state, versus downstream output distributions,  thereby anchoring accumulation rate and behavioral disparity to one another.

Practically, FP32 currently remains the only identified way to escape divergence, although it is expensive.  Other mechanisms are worth investigating: periodic cache refresh (recomputing KV tensors from scratch at fixed intervals), smoothed KV updates (averaging cached and recomputed tensors to reduce drift), or high-precision cache (keep a FP32 cache tensor while keeping every other operation in FP16). All trades compute or memory for less divergence and are probably easier than full FP32 inference on production.

Decision boundary analysis shows that behavioral consequence depends on proximity to the decision boundary, but doesn‘t specify how much accumulated divergence pushes a state across a boundary.  Experiments in adversarial robustness, injecting prespecified noise directly into KV tensors and measuring the flip rate as a function of noise amplitude, which would determine susceptibility thresholds and tell us how close to the decision boundary a model must be for a divergence of a specific magnitude to produce an observed behavioral flip.

In the end, a few more extensions could guarantee generality:  Formal quantifying of the GQA amplification factor across various sharing ratios,  running on instruction-tuned models with extended generation budgets, and characterizing the interplay of KV cache divergence with cache compression schemes (e.g., quantization, eviction, low-rank), where additional precision loss may compound the effects reported here.

\section{Conclusion}
One can assume that, for almost all deployed transformer systems, KV caching is numerically equal to cache-free inference. We have demonstrated that this assumption breaks down with standard FP16 inference: cache-ON and cache-OFF execution paths use different floating-point accumulation order,  resulting in reproducible, deterministic divergence of decoded token sequences for every model, strategy, and input we examined.

We characterized this divergence as a consistent behavior through five experiments. This is a true artifact of the system: all runs diverge, and even with greedy decoding and no sampling for randomness, the system still diverges. This is caused by the non-associativity of FP16, as shown by a perfect falsification: FP32 reduces divergence to a noise floor at least eight orders of magnitude smaller than FP16. Also, it propagates in architecturally predictable ways (sharp amplification in GQA models at first Transformer‘s GQA layers, broadly distributed across all Gemma layers) and cannot be fixed with residual stream interventions, by any number of patched layers. The behavioral importance of the divergence depends on proximity to the decision boundary: examples that are near the decision boundary flip early and have high KL divergence.

These results confirm that FP16 KV cache inference is not deterministically equivalent to cache-free inference,  give a mechanistic explanation based on floating-point arithmetic, and reveal that direct KV cache patching is the intervention needed to restore equivalence. As GQA and MQA rise to become the dominant attention variants in production LLMs, the amplification effect we describe will become more widespread, making the correctness of cached inference a more significant property for system designers.

\bibliographystyle{plain}
\bibliography{ref}

\end{document}